\documentclass[conference]{IEEEtran}

\IEEEoverridecommandlockouts
\usepackage{cite}
\usepackage{amsmath,amssymb,amsfonts}
\usepackage{algorithmic}
\usepackage{graphicx}
\usepackage{textcomp}
\usepackage{subcaption}
\usepackage{xcolor}
\def\BibTeX{{\rm B\kern-.05em{\sc i\kern-.025em b}\kern-.08em
    T\kern-.1667em\lower.7ex\hbox{E}\kern-.125emX}}
\begin{document}

\title{Controllable Path of Destruction}


\author{\IEEEauthorblockN{Matthew Siper}
\IEEEauthorblockA{\textit{Game Innovation Lab} \\
\textit{New York University}\\
New York, USA \\
ms12010@nyu.edu}
\and
\IEEEauthorblockN{Sam Earle}
\IEEEauthorblockA{\textit{Game Innovation Lab} \\
\textit{New York University}\\
New York, USA \\
sam.earle@nyu.edu}
\and
\IEEEauthorblockN{Zehua Jiang}
\IEEEauthorblockA{\textit{Game Innovation Lab} \\
\textit{New York University}\\
New York, USA \\
zehua.jiang@nyu.edu}
\and
\IEEEauthorblockN{Ahmed Khalifa}
\IEEEauthorblockA{\textit{Institute of Digital Games} \\
\textit{University of Malta}\\
Msida, Malta \\
ahmed@akhalifa.com}
\and
\IEEEauthorblockN{Julian Togelius}
\IEEEauthorblockA{\textit{Game Innovation Lab} \\
\textit{New York University}\\
New York, USA \\
julian@togelius.com}
}

\maketitle

\begin{abstract}
Path of Destruction (PoD) is a self-supervised method for learning iterative generators. The core idea is to produce a training set by destroying a set of artifacts, and for each destructive step create a training instance based on the corresponding repair action. A generator trained on this dataset can then generate new artifacts by repairing from arbitrary states. The PoD method is very data-efficient in terms of original training examples and well-suited to functional artifacts composed of categorical data, such as game levels and discrete 3D structures. In this paper, we extend the Path of Destruction method to allow designer control over aspects of the generated artifacts. Controllability is introduced by adding conditional inputs to the state-action pairs that make up the repair trajectories. We test the controllable PoD method in a 2D dungeon setting, as well as in the domain of small 3D Lego cars.
\end{abstract}


\section{Introduction}

Self-supervised learning in various guises has enabled dramatic advances in generative AI over the last decade. Generative Adversarial Networks~\cite{goodfellow2020generative} and diffusion models~\cite{ho2020denoising} have enabled the creation of high-quality images in a variety of styles, and transformers underlie large language models~\cite{wang2019language} which are poised to impact any field of human activity which involves text processing. These deep learning architectures are generally tied to particular representation formats: in particular, transformers operate on sequences of tokens and GANs and diffusion models work with matrices of real numbers which are interpreted as RGB values. But what these methods are above all is data-hungry. A GAN or diffusion model is trained on at least thousands, often millions, of images~\cite{baio2022exploring}, and LLMs are typically trained on terabytes of text~\cite{openai2023gpt}.

This leaves a large space of creative domains which are not naturally expressed as matrices of real numbers or sequences of tokens underserved by current self-supervised learning methods. Even more importantly, for most domains comparatively little data is available to train on, rendering existing self-supervised methods largely ineffective. 

Path of Destruction is a recently developed self-supervised method for learning generators of structured content. The method is very data-effective. Although it was developed separately from diffusion models, it has conceptual similarities. The basic idea is to iteratively destroy the target content, one small change at a time, and create a dataset of the changes. In the generated dataset, each instance has part of the destroyed state as an input and the associated repair action (reverse of the destruction action) as the target. By rolling out multiple paths of destruction, an arbitrarily large set of repair actions can be created even when starting with a small initial set of artifacts. Standard supervised learning can then be used to train a generator that generates novel artifacts by ``repairing'' from random starting states.

\begin{figure}[t]
    \centering
    \includegraphics[width=0.9\linewidth]{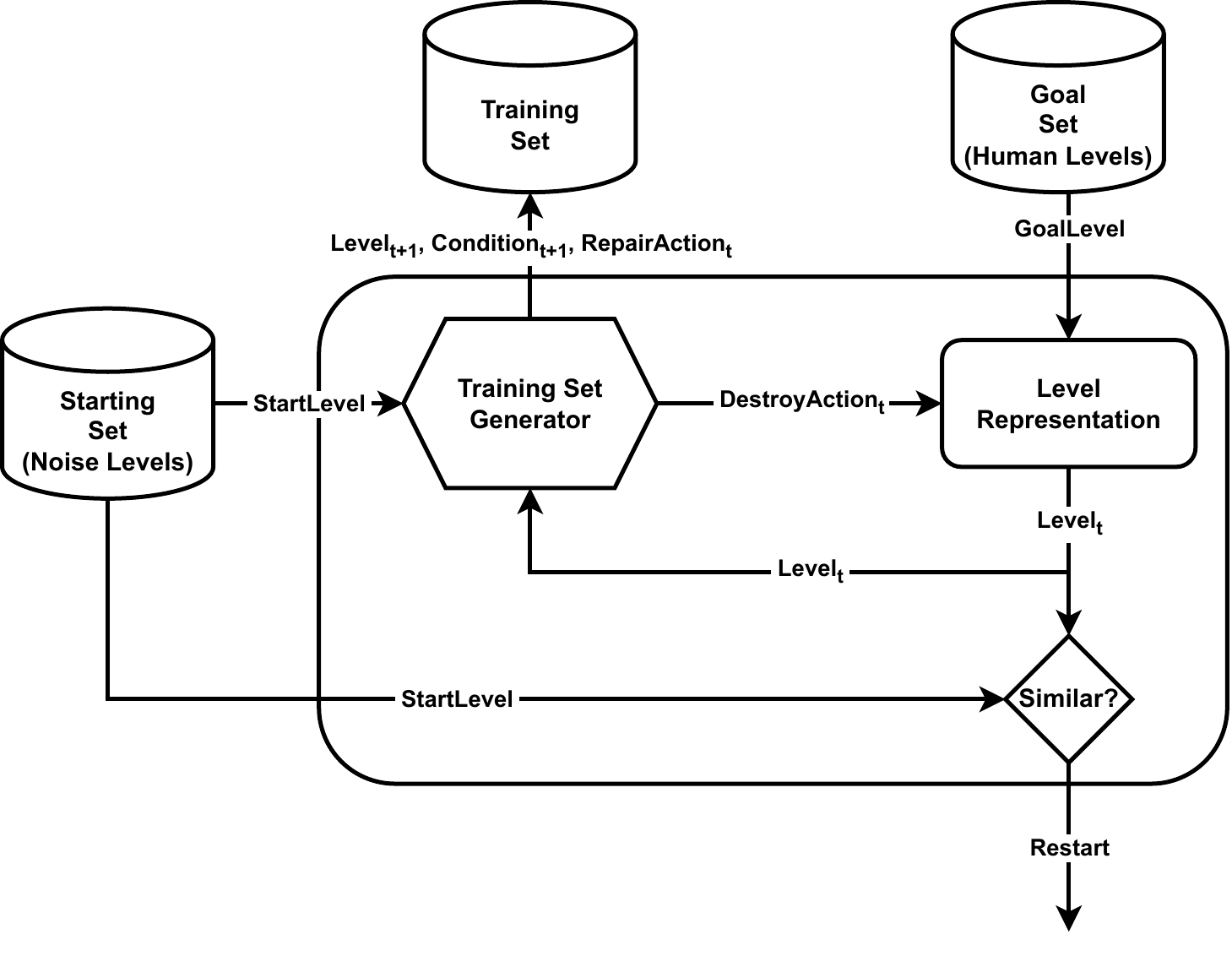}
    \caption{System diagram of the path of destruction data generation loop including the new condition signal.}
    \label{fig:pod_gen}
\end{figure}

The original Path of Destruction method~\cite{siper2022path} could generate novel, playable levels for several 2D games based on training on as little as 5 source levels, although performance improved with more levels to train on. However, that algorithm provided no way for the user to influence the design of the generated content beyond curating the training set. In this paper, we describe and evaluate a new version of the Path of Destruction algorithm which allows a user to control aspects of the output. This opens up new avenues for using Path of Destruction as a design tool in interactive applications~\cite{yannakakis2014mixed}.

The basic idea of Controllable Path of Destruction is to include conditional inputs in the generator. These conditional inputs are also part of the training data that is generated by destroying the artifact and are based on the properties of the artifact that was originally destroyed in the process. For example, if the controllable aspect of the generator is the number of tiles of type X, then the dataset will contain a conditional input for the corresponding tile count. All instances deriving from the destruction of a particular artifact will have the tile count from that artifact. 

To improve the performance of our algorithm, in particular, for settings with limited training data, we propose signed inputs, where the conditional input is not the absolute value of a feature, but whether the artifact that is being generated is currently above or below the target value. Using signed inputs (positive, negative, zero) helps to have a bigger distribution over these different conditions compared to using direct values. We hypothesize that this helps the network to learn the meaning of these conditions better than using actual values.

We test our algorithm in two domains: 2D ``Zelda'' levels (single-room levels for a much-simplified version of the dungeon system of The Legend of Zelda (Nintendo, 1986)) and small 3D Lego cars.

\section{Related work}

The Path of Destruction method is conceptually and functionally related to several other methods originating from different communities. One of these closely related methods is WaveFunctionCollapse (WFC), a method that has origins in the graphics community but was popularized by Maxim Gumin~\cite{gumin2016wfc}. While WFC is  used for PCG in some games, in particular indie games~\cite{thompson2022townscaper}, it can also be seen as a form of constraint satisfaction and related to constraint satisfaction research in the AI community~\cite{karth2017wavefunctioncollapse}. WFC operates on (preferably low-dimensional) pixel- or tile-based 2D artifacts and learns to generate new artifacts in the same style. It does this by learning the local rules that relate the pixels/tiles in the input artifact, and ensuring that the generated artifact adheres to the same rule. Every patch in the generated artifact must have occurred somewhere in the input. Like WFC, Path of Destruction~\cite{siper2022path} largely learns local rules and is highly data-efficient, but uses a deep neural network and tries also to learn global rules as the training data is generated towards certain targets.

Another highly influential family of methods with conceptual similarities to Path of Destruction is diffusion models\footnote{Path of Destruction was developed independently of diffusion models, and the relation was pointed out to the authors of the first Path of Destruction paper after it had been posted on ArXiv}~\cite{sohl2015deep,ho2020denoising}. Diffusion models are typically trained on large sets of images~\cite{baio2022exploring}. On each image, noise is iteratively applied until only noise remains. The model is then iteratively trained to regenerate the original image from the noisy version of the image. The Path of Destruction is conceptually similar in that it also teaches neural networks to recreate artifacts after destroying them. However, whereas diffusion models operate in parallel (on the whole image at a time) and with continuous-valued pixels, Path of Destruction operates sequentially, changing one pixel/tile/element at a time. The current implementation operates on discrete, categorical tile values which also differentiates it from diffusion models. 
Given a sufficiently wide definition of ``diffusion models'', one might indeed choose to see Path of Destruction as a type of diffusion model; we have no strong opinion on this matter. However, we note that the data efficiency of Path of Destruction is much better than any diffusion model we know of, in the sense that it can learn to generate good artifacts with very few training examples. We also point out that PoD was able to learn functional relations between tiles that enabled it to generate playable levels which is not the case for diffusion models. 

A third method that should be mentioned in this context is procedural content generation via reinforcement learning, or PCGRL for short~\cite{khalifa2020pcgrl}. In PCGRL, reinforcement learning algorithms are used to train neural nets to construct levels. Path of Destruction shares its scanline-like mode of operation with the ``narrow'' variety of PCGRL, and the controllable version of Path of Destruction presented here is inspired by Controllable PCGRL~\cite{earle2021learning}. As discussed later, the very same network could plausibly be trained by either method, suggesting interesting combinations of these methods. However, unlike PCGRL where a reward function is used to train the generator network, Path of Destruction is a self-supervised method trained on existing artifacts.

In this paper, we apply Controllable Path of Destruction to game-level generation, a problem that has been widely studied within procedural content generation (PCG) research~\cite{shaker2016procedural}. Typically, PCG is motivated by giving game developers new tools, but procedurally generated levels are also important for open-ended learning~\cite{risi2020increasing}. In recent years, machine learning techniques have increasingly been applied to PCG, including self-supervised methods~\cite{summerville2018procedural,liu2021deep,guzdial2022procedural}.

\begin{figure}[t]
    \centering
    \includegraphics[width=0.65\linewidth]{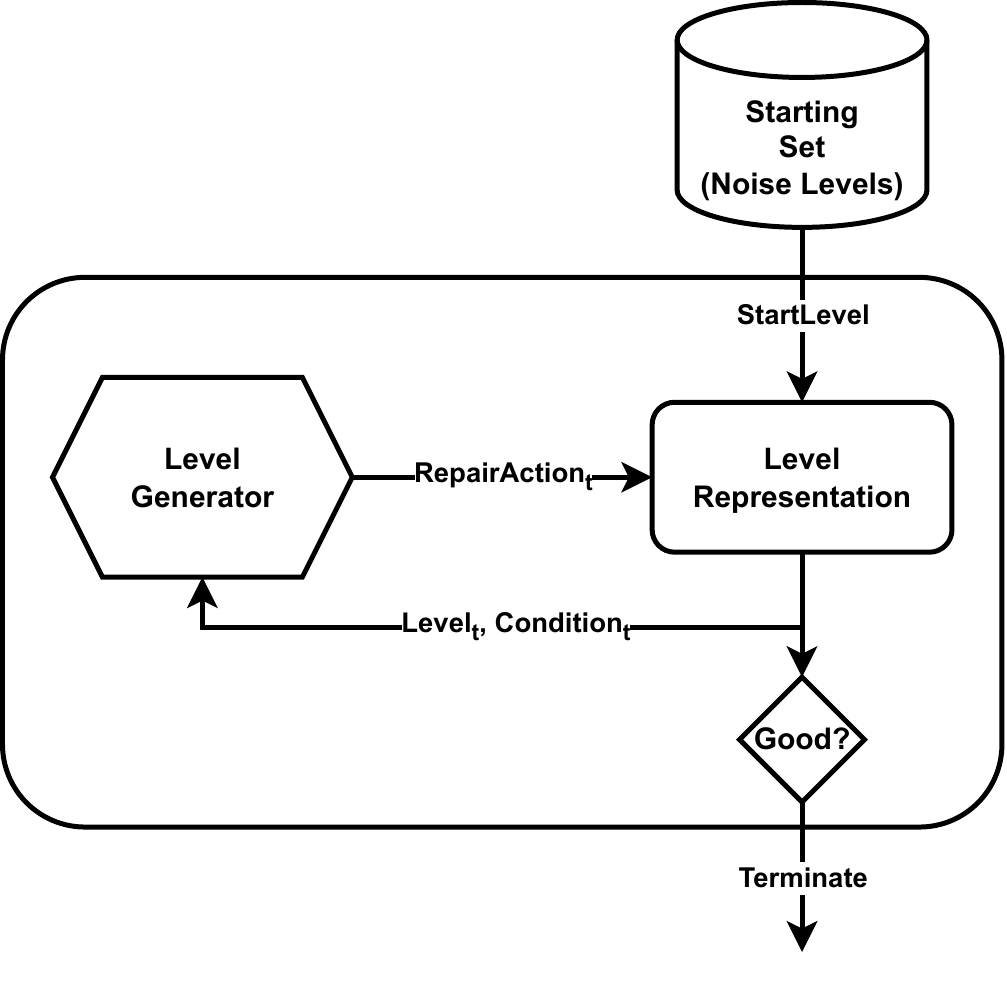}
    \caption{System diagram of the inference step using the trained model from the path of destruction. The difference between this figure and the normal path of destruction is the condition signal.}
    \label{fig:pod_infer}
\end{figure}

\begin{figure*}[t]
    \centering
    \begin{subfigure}[t]{0.4\linewidth}
        \centering
        \includegraphics[width=.48\linewidth]{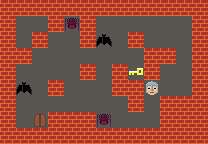}
        \includegraphics[width=.48\linewidth]{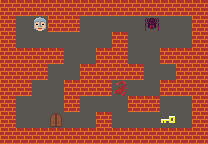}
        \includegraphics[width=.48\linewidth]{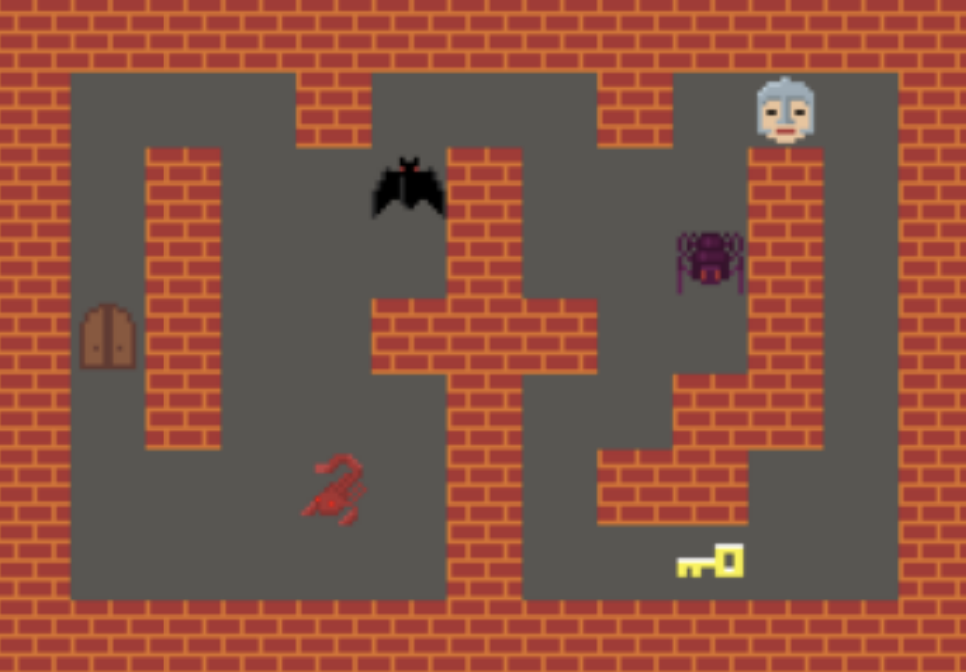}
        \includegraphics[width=.48\linewidth]{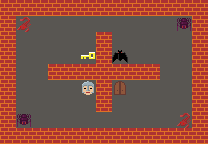}
        \caption{Zelda}
        \label{fig:zelda_domain}
    \end{subfigure}
    \begin{subfigure}[t]{0.5\linewidth}
        \centering
        \includegraphics[width=.48\linewidth]{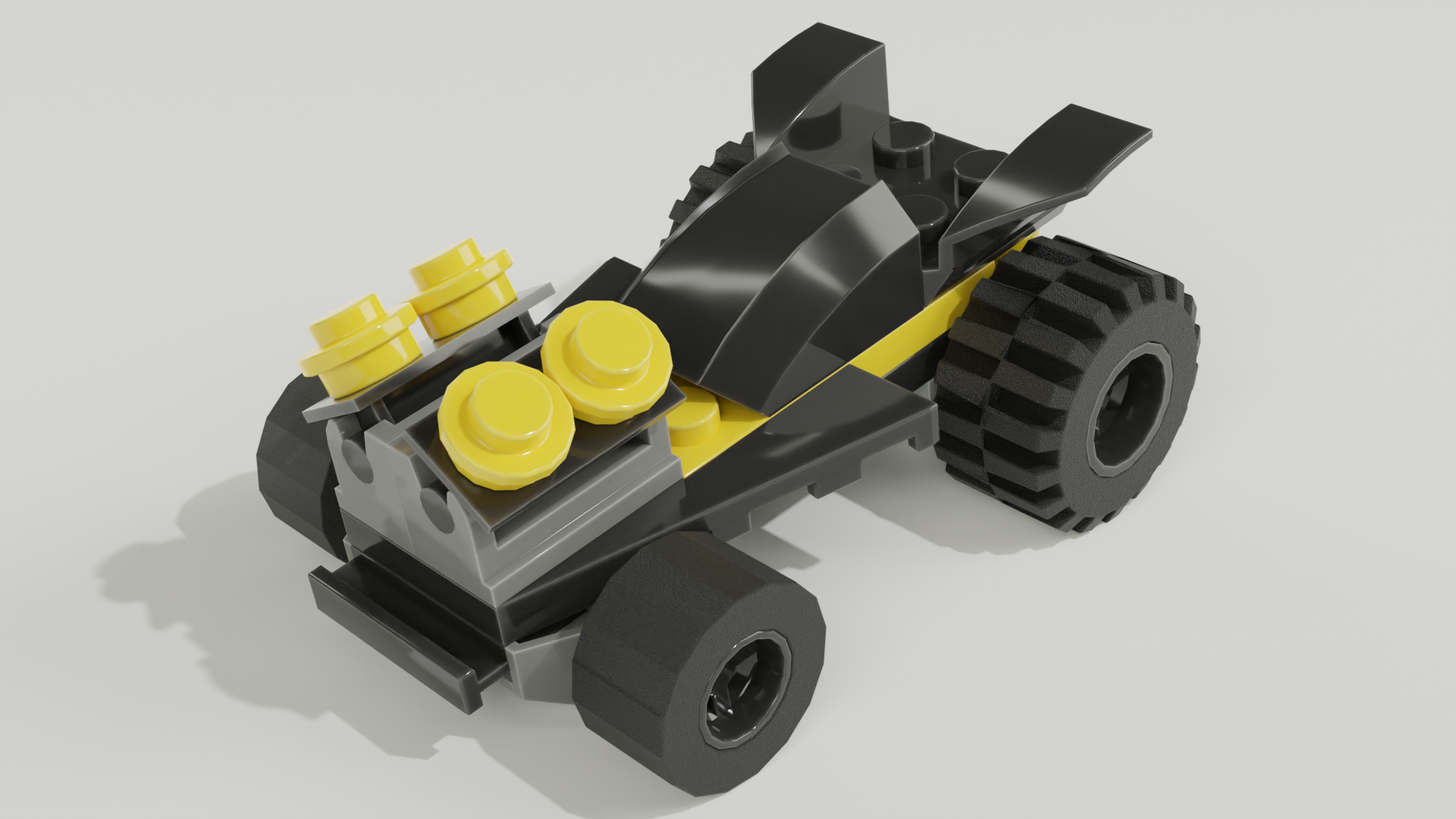}
        \includegraphics[width=.48\linewidth]{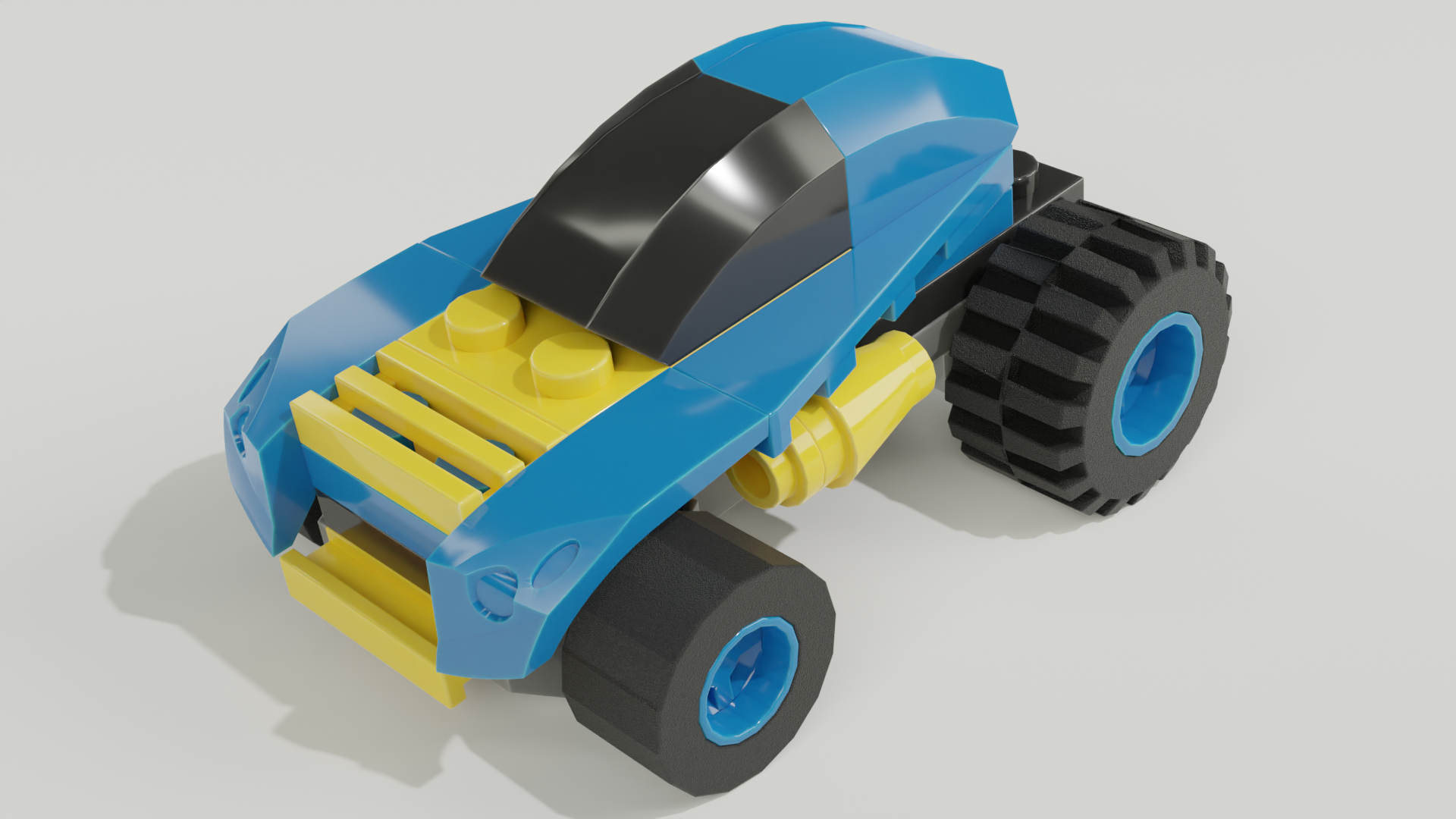}
        \includegraphics[width=.48\linewidth]{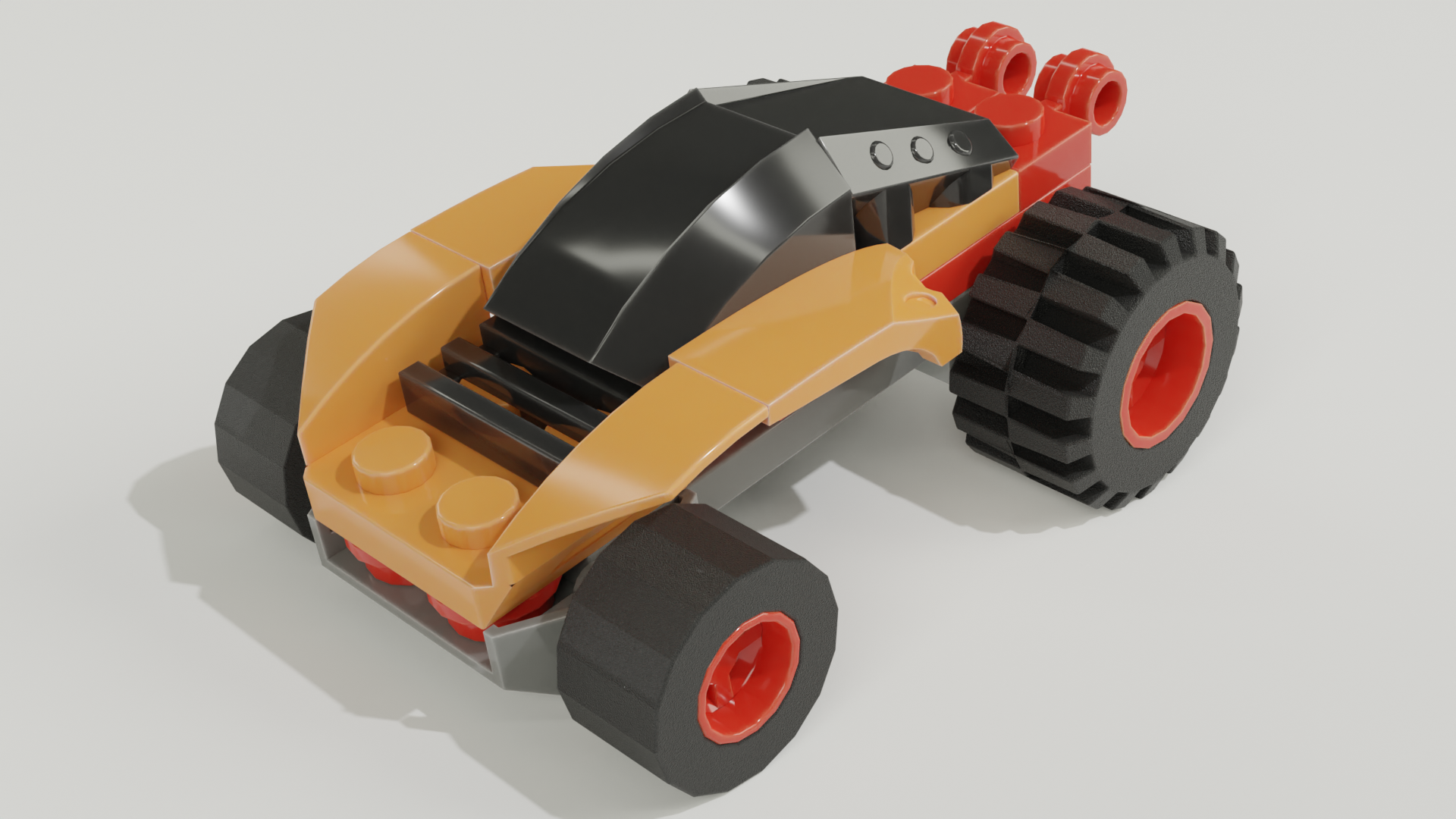}
        \includegraphics[width=.48\linewidth]{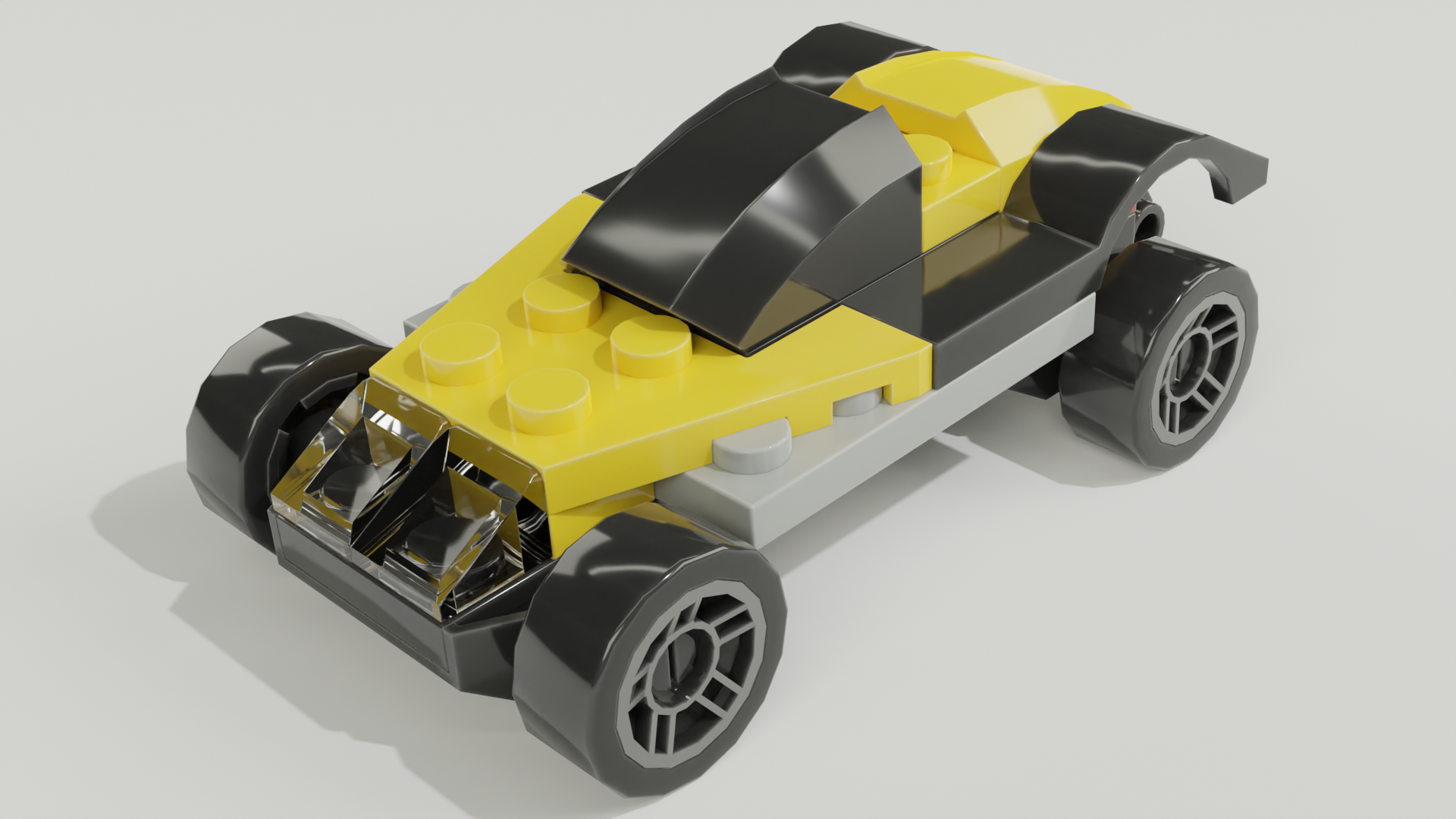}
        \caption{Lego Car}
        \label{fig:lego_domain}
    \end{subfigure}
    \caption{Examples from the goal set for both domains.}
    \label{fig:domains}
\end{figure*}

\section{Controllable Path of Destruction}
Path of Destruction is a data augmentation method introduced by Siper et al.~\cite{siper2022path} that helps to create a big training dataset given a small number of training data. The method is similar to diffusion models~\cite{ho2020denoising} but it can work with discrete data and doesn't have any conditions on the training noise. Furthermore, Path of Destruction can work using a very small amount of data. To avoid confusion, the input data that is used to create the training data is called the goal set, while the output of the technique is called the training set.

The original Path of Destruction follows the following steps to create the training set:
\begin{enumerate}
    \item Pick a random level/object from the goal set and set the current state ($Level_t$) equal to it.
    \item Pick a random location on the current state.\label{step:randlvl}
    \item Change the current state with a randomly selected value ($DestroyAction_t$).
    \item Record the value of the current location before the change ($RepairAction_t$) and the new state ($Level_{t+1}$) as a state-action pair to save it in the training set.\label{step:gen_condition}
    \item If the new state doesn't follow a certain noise distribution called Starting set, Go back to step~\ref{step:randlvl}.
    \item Repeat the above steps until we have a big enough number of state action pairs.
\end{enumerate}

After the training set is generated, we train a neural network using supervised learning to learn how to reverse the destruction. After the training is done, the network can be used to generate new content by following the following steps:
\begin{enumerate}
    \item Generate a random state ($StartLevel$) that follows the same distribution of the starting set and set it as the current state ($Level_t$).\label{step:infer_condition}
    \item Pick a random location of the state and apply the network using the current state ($Level_t$).
    \item Replace the location value with the output from the network ($RepairAction_t$) to get the new state ($Level_{t+1}$).
    \item Repeat the above steps until the object reaches a certain minimum quality (for game levels until the level is playable) or time is up.
\end{enumerate}

One of the main issues of this method is that it is not easy to control any of the features of the generated content. The only way of control is through the starting state which doesn't give much understanding about the final generated content. To battle this problem, we took inspiration from Earle et al.'s work~\cite{earle2021learning} and decided to explore the idea of having a control signal that can influence the algorithm toward generating content with specific features. For example, if we are generating a level for Super Mario Bros (Nintendo, 1985), there could be an input signal that controls how many jumps the player needs to do to beat the level. Figure~\ref{fig:pod_gen} and \ref{fig:pod_infer} show the changes to the flow explained above. The big difference is during generating the data set we need to make sure to record the target condition value ($Condition_t$) at step~\ref{step:gen_condition} as part of the training data. For the inference side, the user needs to pick the target condition ($Condition_t$) when generating the starting state at step~\ref{step:infer_condition}.

Originally these input signals were absolute target values of what is required. For example, if we want to have 5 enemies, we need to give that value to the model. The problem was there is few goal states which made the number of conditions limited and doesn't cover the whole space of conditions. For example, if we have 5 goal levels, we only have 5 different conditions and some of them might be repeated. In early experiments, the learned model was not able to generalize and understand other values outside the ones that it was trained on. We decided to use the same idea from Earle et al. work~\cite{earle2021learning} where instead of using the target values, we use a signed relative signal that tells the system if it should increase (1), decrease (-1), or keep (0) that target value. For example, if we are trying to generate a level with 5 enemies, we first check the current level number of enemies if there are 8 enemies then the signal is -1.

We updated the main algorithm to incorporate this new condition signal. For the data generation process, we now calculate the signal sign by calculating the current value of the condition and comparing it with the goal state value and then get added as part of the input state in the training data. The training data now is observation ($Level_t$), condition signal ($Condition_t$), and output ($RepairAction_t$) where both the observation and condition signal are the input for the machine learning model. For the generation phase, we need to calculate at every step the value of the condition with respect to the user target value. For example, if the user wants a level with 20 enemies and the current state has 40, then the condition is equal to -1.

Using the signed relative signal helps the network to learn a more generic meaning of the signal than targets as we have more examples for it. In the future, we could use a more intelligent way of destruction that make sure that all the different conditions are well covered. As in the current method, some conditions might be harder to cover. For example, imagine one of the conditions is the longest path in a maze, most of the destruction will be decreasing that path which makes the condition always 1 in most of the training data. This problem becomes more apparent when there the number of conditions is increased where certain combinations might rarely appear during random destruction.

\section{Experiments}

To analyze and study this controllable model, we decided to test the model on two domains (figure~\ref{fig:domains}):
\begin{itemize}
    \item \textbf{Zelda Level Generation}: is a 2D top-down demake of the dungeon system of The Legend of Zelda (Nintendo, 1986). The goal of the game is to get a key and reach the door while avoiding being killed by enemies. The player can kill enemies using their swords. The goal is to create a playable level with one key, one door, and one player.
    \item \textbf{Lego Car Generation}: is a 3D framework using a constrained set of Lego pieces to create a 3D car that is functional (has 4 wheels).
\end{itemize}
We picked these two domains as they show the diversity and the ability of the method to work between a tile-based restrictive level to more free-form Lego pieces.

\begin{figure}[t]
    \centering
    \includegraphics[width=0.9\linewidth]{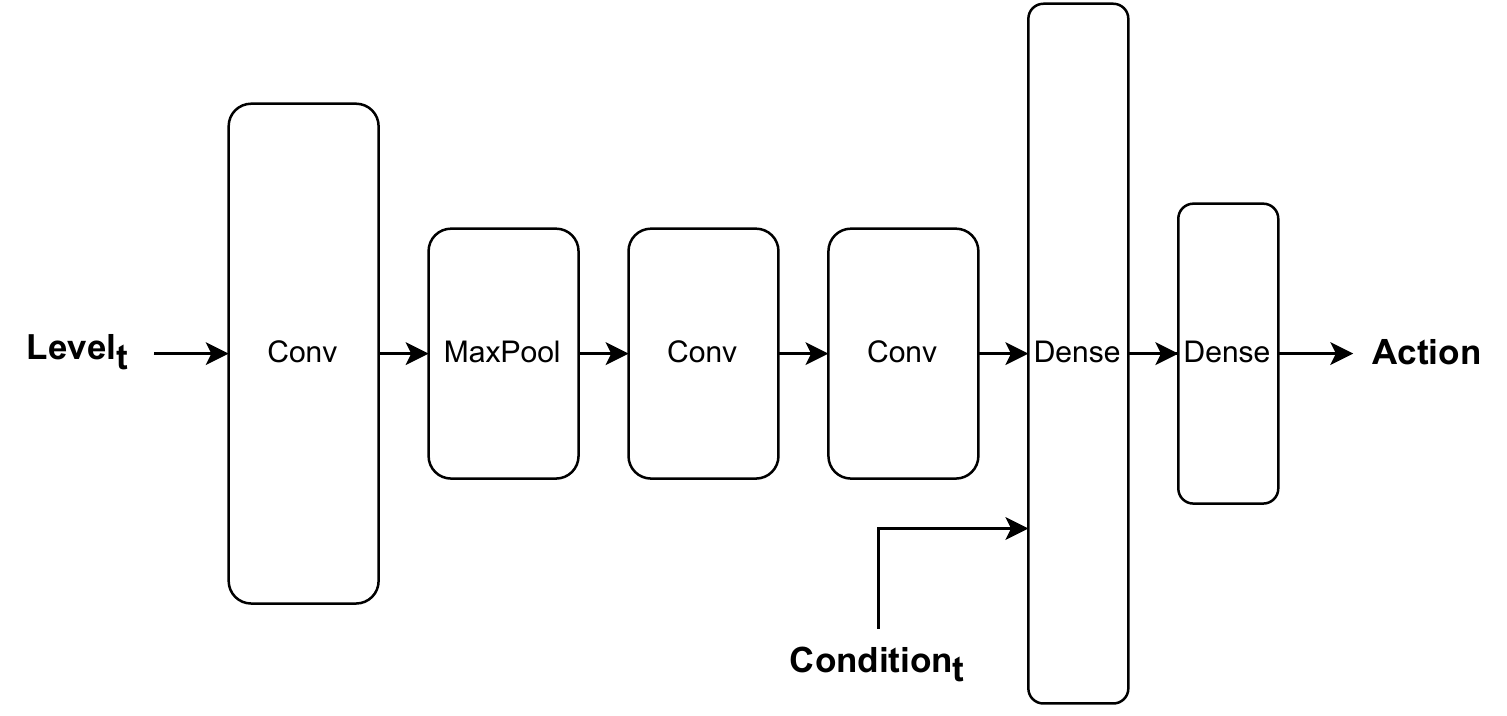}
    \caption{The network architecture used as the generation model.}
    \label{fig:architecture}
\end{figure}

For both problems, we used a simple architecture (figure~\ref{fig:architecture}) similar to the Deep Q-Learning architecture~\cite{mnih2013playing}. The architecture consists of 3 convolution layers followed by two fully connected layers. All the intermediate layers are using a Relu activation function, while the final layer is using a Softmax layer. We also added a MaxPooling layer after the first layer to reduce the size of the input state. The network takes two inputs: the current observation ($Level_t$) and the current condition signal ($Condition_t$). Both observation and condition signal is encoded as one-hot encoding as they are discrete values. The current observation goes through the convolution layers and gets concatenated to the condition signal for the last two layers. The difference between the architecture of Lego and Zelda problems is that in Lego the convolution and max pooling layers are 3D instead of 2D.

\subsection{Zelda Level Generation}
For this problem, we cared about 3 different controllable parameters:
\begin{itemize}
    \item \textbf{Number of Enemies:} The number of enemies in the generated level.
    \item \textbf{Distance to Nearest Enemy:} The Manhattan distance to the nearest enemy in the generated level.
    \item \textbf{Solution Length:} The number of player steps that are needed to solve the current level (The Manhattan distance between the player to the key added to the Manhattan distance from the key to the door).
\end{itemize}
We picked these parameters as they control the target experience of the game. For example, having a higher number of enemies will make the game harder and has more action moments while having closer enemies makes the game more intense and reactive when you start. Finally, the solution length controls how long should the game takes to solve and also control if the player passes through multiple enemies.

For this experiment, we used full observation instead of partial observation which was used before~\cite{siper2022path}. The reason for that is we wanted the network to have the ability to learn easily about these conditions. The observation size is 19x19 where the extra outside areas were padded using solid tiles. The starting set distribution follows the distribution of Zelda's goal-set levels. We used 50 goal levels (figure~\ref{fig:zelda_domain}) to generate 962,500 training data. We train three different networks for 250 epochs with a batch size of 256 using Adam optimizer with default settings.

For the generation process, we sampled a random level from the distribution and run the model sequentially on all the locations for 105 steps or until we find a playable level within a threshold for the condition target. A level is playable if it is fully connected with one key, door, and player. Table~\ref{tab:zelda_parameters} shows the range of the conditional values and the different conditional thresholds that were used in the generation process. All the values were picked relative to the training data. We generated 500 levels using the network with condition values covering the space of different target values.

\begin{figure*}[t]
    \centering
    \begin{subfigure}[t]{.49\linewidth}
        \centering
        \captionsetup{width=.9\linewidth}
        \includegraphics[width=.49\textwidth]{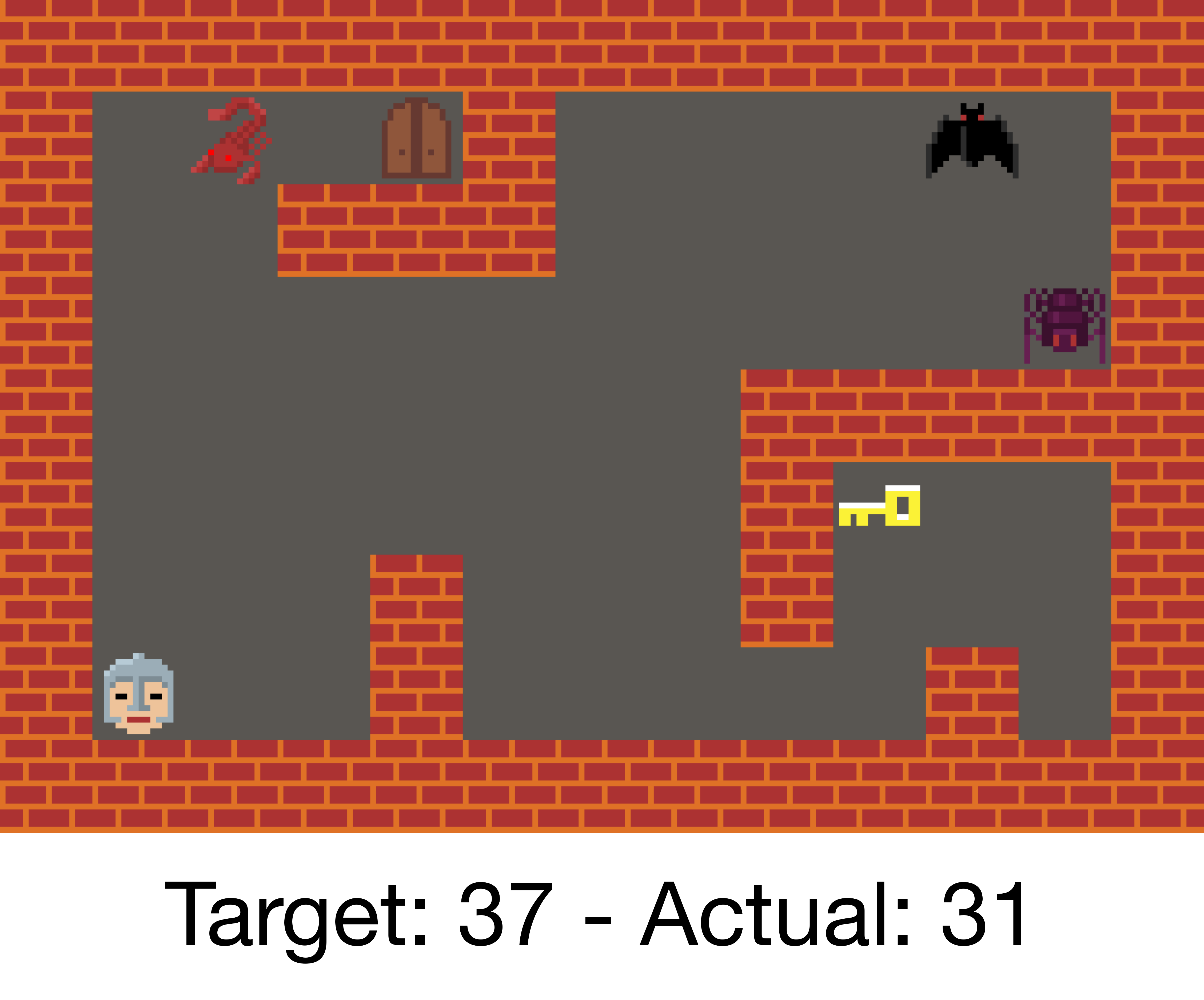}
        \includegraphics[width=.49\textwidth]{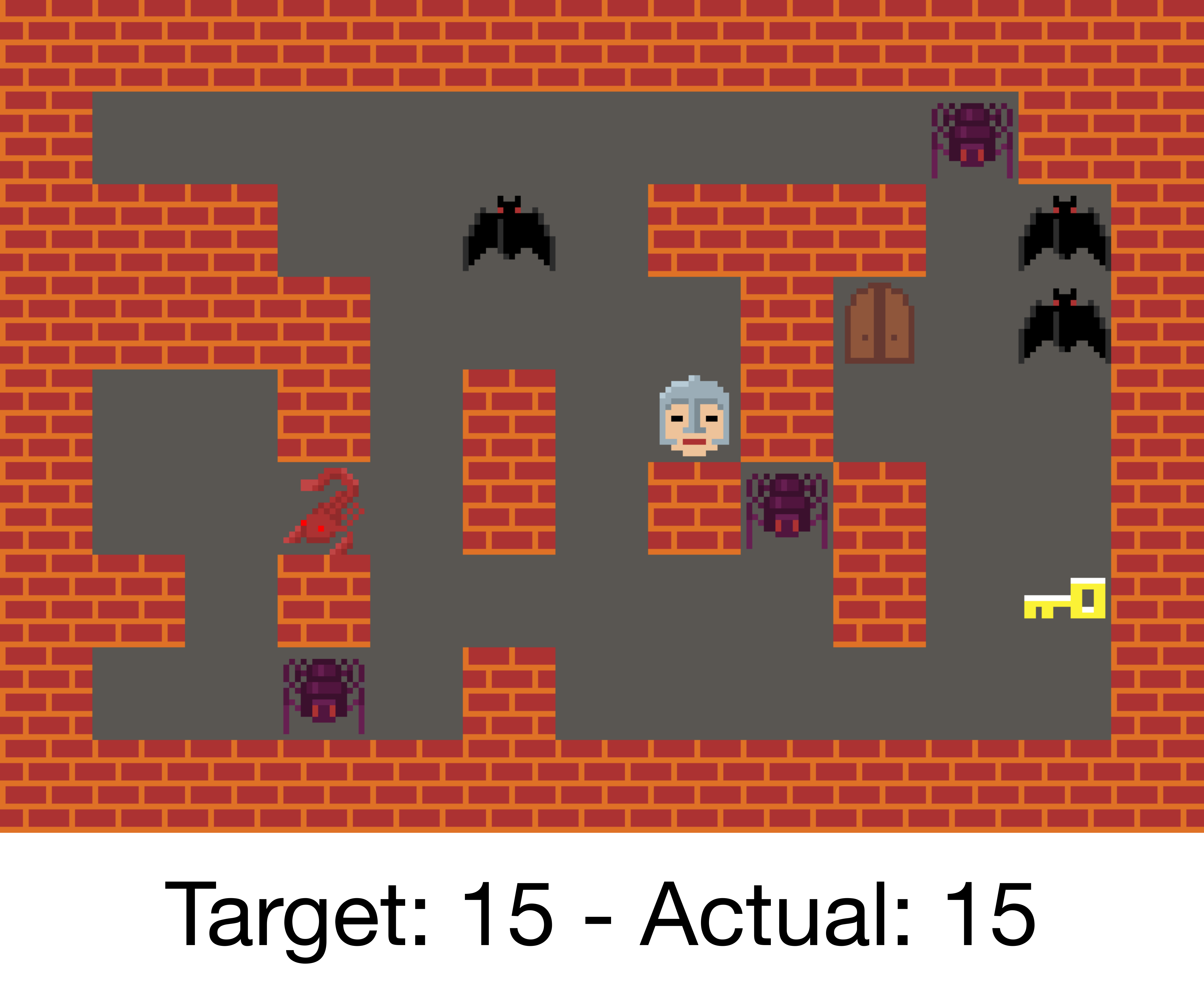}
        \caption{Zelda levels generated by prompting the model for high/low path length.}
        \label{fig:zelda_ex_path}
    \end{subfigure}
    \begin{subfigure}[t]{.49\linewidth}
        \centering
        \captionsetup{width=.9\linewidth}
        \includegraphics[width=.49\textwidth]{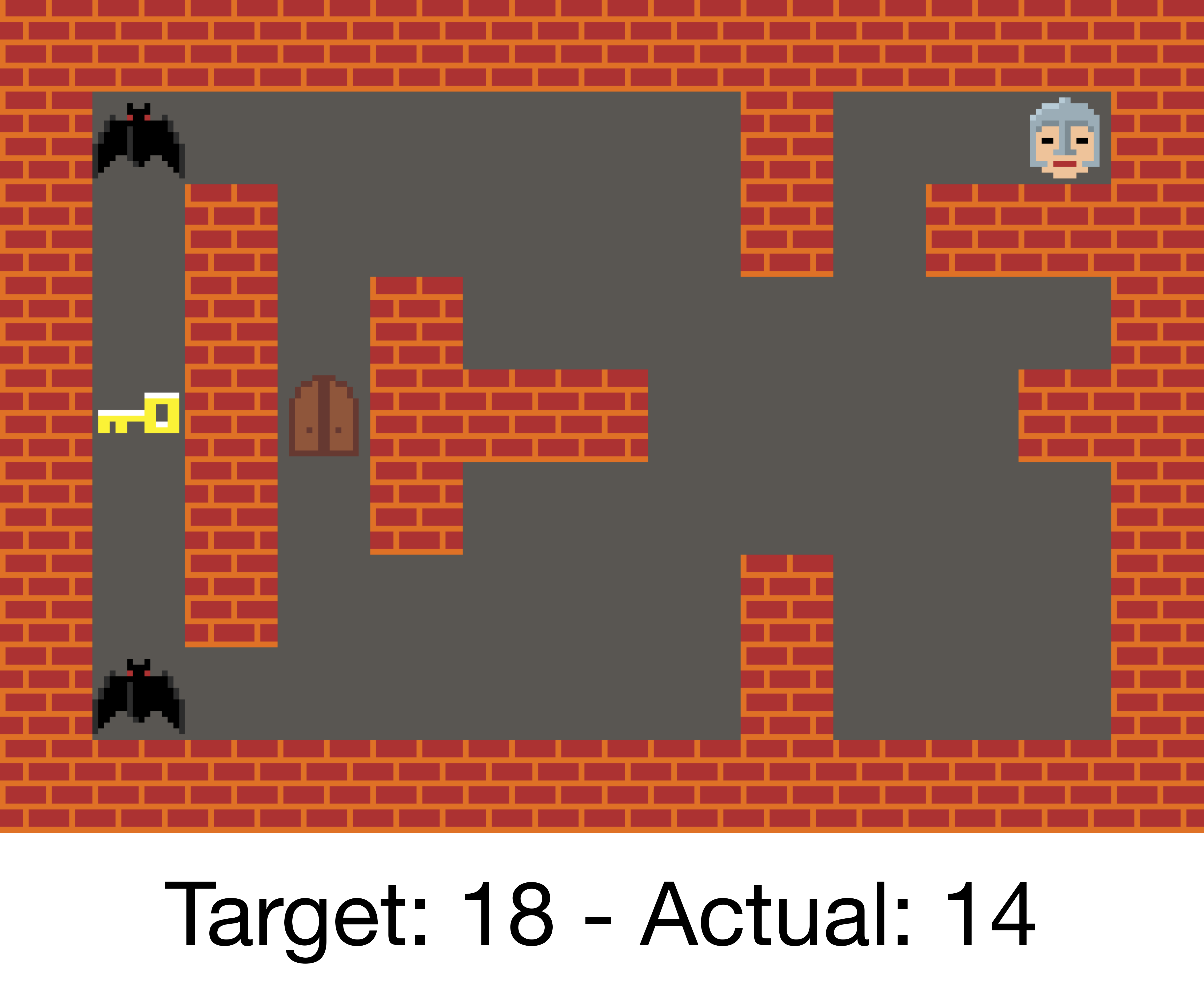}
        \includegraphics[width=.49\textwidth]{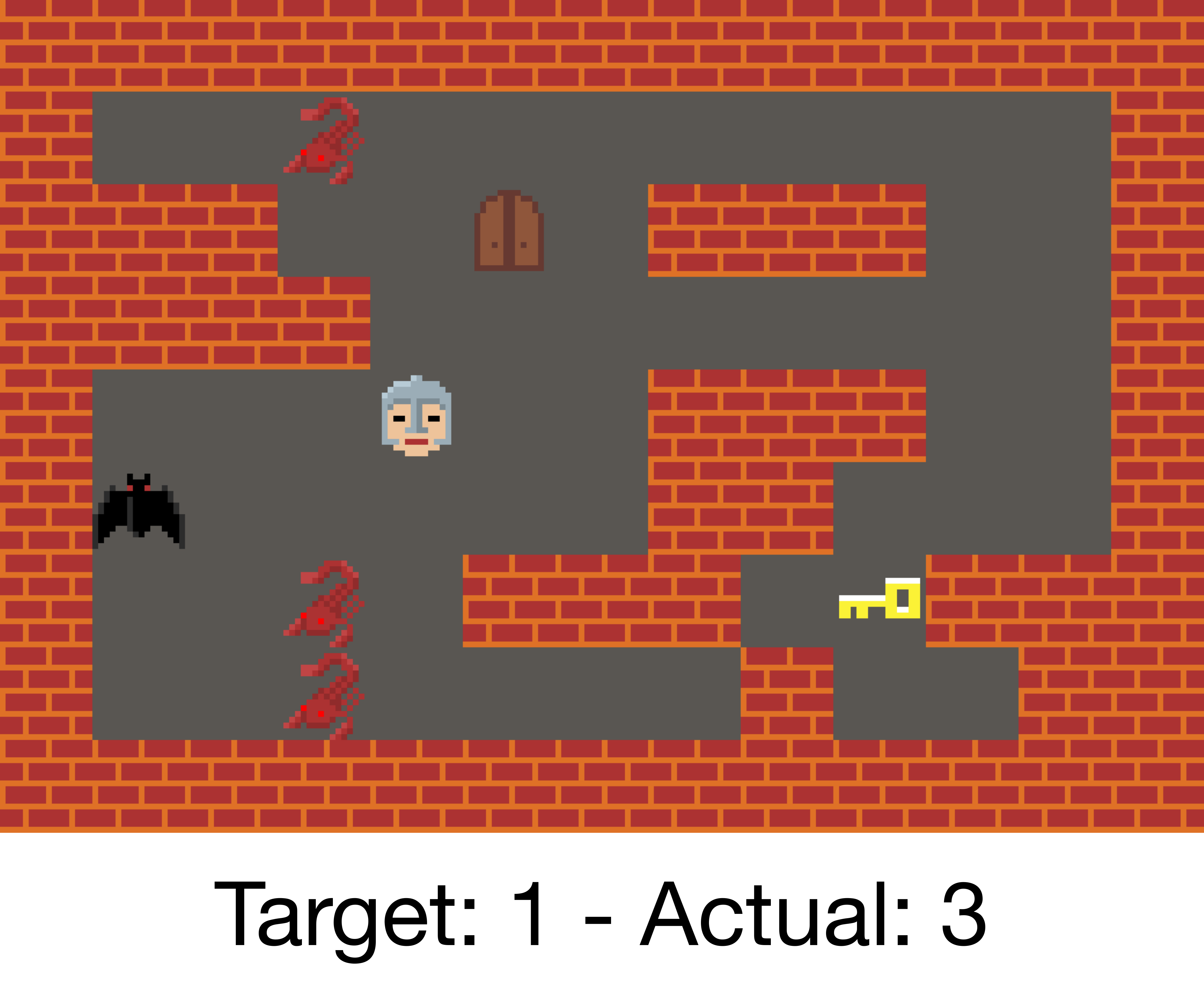}
        \caption{Zelda levels generated by prompting the model for high/low distance between the player and the nearest enemy.}
        \label{fig:zelda_ex_enemies}
    \end{subfigure}
    \caption{Generated Zelda levels using one of the trained neural networks.}
    \label{fig:zelda_examples}
\end{figure*}

\begin{table}[t]
    \centering
    \begin{tabular}{|l||c|c|c|}
        \hline
       Condition  & Min Value & Max Value & Threshold \\
       \hline
       \hline
       Number of Enemies  & 0 & 5 & 4 \\
       \hline
       Distance to Nearest Enemy & 0 & 12 & 5 \\
       \hline
       Solution Length & 10 & 31 & 6 \\
       \hline
    \end{tabular}
    \caption{The conditions values parameters used for the Zelda level generation.}
    \label{tab:zelda_parameters}
\end{table}

\subsection{Lego Car Generation}

For the Lego car generation, we use the number of blocks as the control parameter. This parameter is simple but it is the most useful one when creating Lego objects as this is a common real-world restriction. In this experiment, we didn't differentiate between the different block types. In the future, we think it makes more sense if the control was distributed on the different types of blocks so people can tell the system the different numbers of blocks they have from each.

In this experiment, we use 37 different block types and an observation size of $6\times6\times6$. Therefore, the state space has size $6\times6\times6\times37$ since we one-hot encode the blocks. There is no restriction about where blocks can and can't go, however, we would like to restrict this in the future to avoid floating and overlapping blocks in the output. The goal set of Lego cars used to generate the training trajectories consists of 15 different cars of varying numbers of component blocks (figure~\ref{fig:lego_domain}). The training set consists of 300,000 state-action pairs produced from continuously iteratively destroying a goal car to a noisy state. We trained three networks for 40 epochs, with a batch size of 256. We used the Adam optimizer with default parameters.

For inference, we generated 520 cars from random starting states where each voxel in the state space is filled with a uniformly randomly sampled block from the 37 possible block types. Each episode consisted of 54 steps. There is no early stopping condition similar to the Zelda generation problem as it is harder to define one for Lego. One might say to stop when there are 4 wheels, the problem with that is the generation could early terminate as the generator might start by placing the wheels first. For the condition parameter, it was tested for values between 15 and 27 blocks. Moreover, to evaluate agent performance during inference, we define two output states of varying success. A ``success" output is one which has four wheels and where the difference between the number of blocks in the output and the target input is no greater than 6. A ``partial success" lacks the difference condition but still contains four wheels (it should be noted that we count the existence of a hubcap without a tire as a wheel). 

\section{Results}
\subsection{Zelda Level Generation}

\begin{figure}
    \centering
    \includegraphics[width=\linewidth]{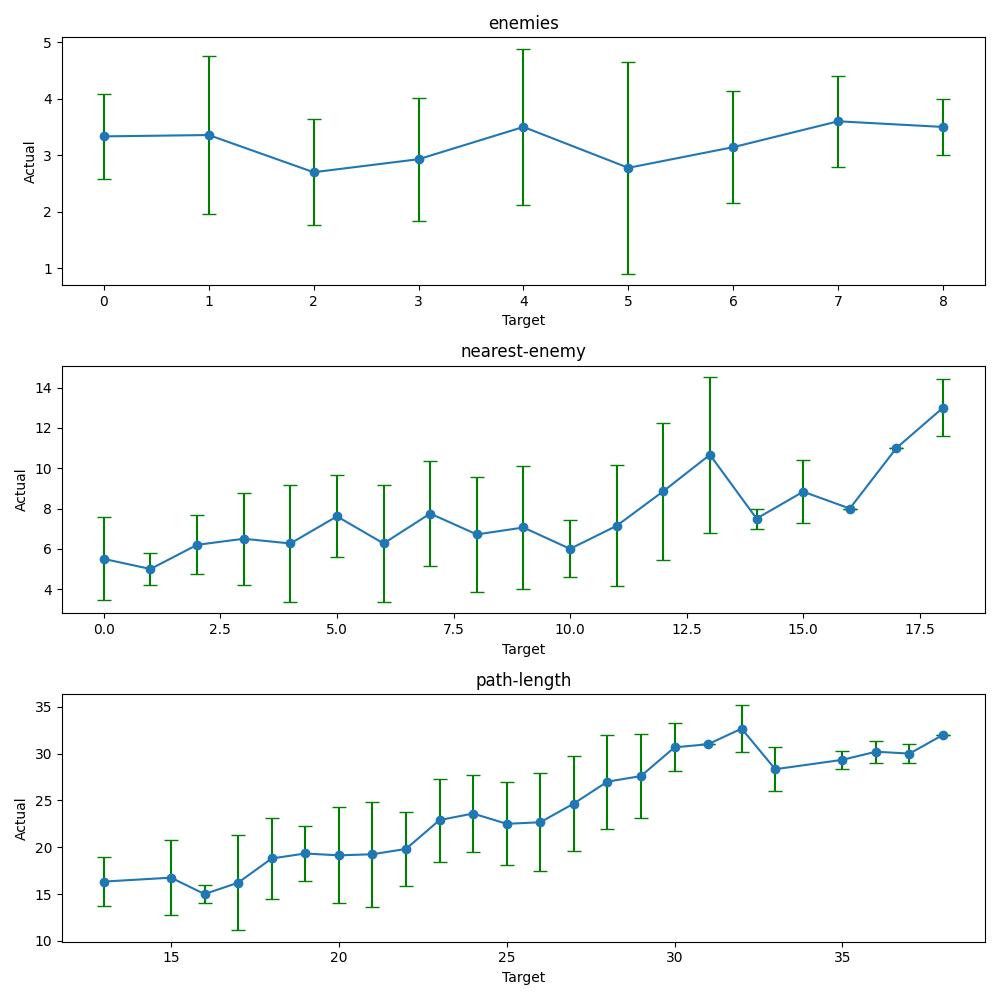}
    \caption{Controllability of Zelda levels. The model learns to effectively control the distance between the player and the nearest enemy, and the path length from player to key to door, but is unable to exert control over the number of enemies in a level, likely because of a lack of levels with diverse numbers of enemies in the training set.}
    \label{fig:zelda_ctrl_metrics}
\end{figure}

Figure~\ref{fig:zelda_ctrl_metrics} shows the controllability of the Zelda agent across the 3 metrics measured: number of enemies, nearest enemy, and path length. The metric with the least amount of controllability was the number of enemies. We attributed this to the fact that the number of enemies from the goal maps was highly concentrated duplicate values. The range of the enemies was also small (from 0 to 7). We think that having a smart destructive agent that creates the dataset could help solve this problem. By contrast, the nearest-enemy and path-length metrics show effective controllability as is evident by their respective curves trending up and to the right. Figure~\ref{fig:zelda_examples} shows some of the generated Zelda levels where the generated level reached the target value for both path length and distance to the nearest enemy.

To test the quality of the generated content, we had to analyze the generated levels in terms of quality and diversity. The quality is simple, we just check the number of playable generated levels. Our generated levels have $78.86 \pm 4.95\%$ playable levels. On the other hand, diversity is a little bit more complex as we need to calculate two values: Inter-diversity and Intra-diversity. Inter-diversity is responsible for calculating how much of the generated levels are different from the goal-set levels, while Intra-diversity makes sure that the generated maps are different from each other. The diversity calculation used is the same one from Siper et al. work~\cite{siper2022path} where we consider two maps to be different if they are $10\%$ different from each other in terms of Hamming Distance. For inter-diversity, the generated output maps are compared to their most similar goal map counterparts. Generated levels that are less than $10\%$ different than their nearest goal map are removed (i.e. considered as duplicates). Our Zelda level generators produce $25.32 \pm 2.3\%$ unique levels by this inter-diversity metric. Intra-diversity is computed by comparing each generated map to the other most similar generated map and removing those that are less than $10\%$ different. $35.40 \pm 1.12\%$ of our generated Zelda levels are unique in terms of intra-diversity. Finally, to compute the total diversity, we first compute the Inter-diversity followed by the intra-diversity; that is, starting with the set of all generated levels, we first remove duplicates w.r.t. goal levels, then remove duplicates w.r.t. other generated levels. The final diversity score (i.e. the percentage of overall unique generated levels) is $24.42 \pm 1.93\%$. Overall, the agent performs reasonably well via being able to produce $78.86 \pm 4.95\%$ playable maps while maintaining a diversity of almost $24.42\%$ and effective controllability with respect to the path length and nearest enemy metrics. 

\begin{table}
    \centering
    \begin{tabular}{|l||c|c|c|}
        \hline
       Model & Playability & Duplicated levels  \\
       \hline
       CESAGAN  & 47.00\% & 60.30\%  \\
       \hline
       PoD & 37.88\% & 0.00\%\\
       \hline
       Controllable PoD & 78.96\% & 26.90\%   \\
       \hline
    \end{tabular}
    \caption{The agent performance w.r.t. its ability to generate diverse and playable maps.}
    \label{tab:zelda_benchmark}
\end{table}

Table~\ref{tab:zelda_benchmark}, we compare the playability and percentage of duplicated levels to the CESAGAN model from ~\cite{torrado2020bootstrapping} and the original Path of Destruction~\cite{siper2022path}. As is evident CESAGAN achieves a modest advantage over our model but this comes at the expense of generating mostly duplicate output. This is compared to controllable PoD which produces mostly new and diverse maps while having the added feature of being effectively controllable on both path length and enemies metrics. On the other hand, the normal PoD was able to have a lot less duplicate levels with almost the same playability levels but in return, it is pretty hard to control the output of the network.

\subsection{Lego Car Generation}

\begin{figure}
    \centering
    \includegraphics[width=\linewidth]{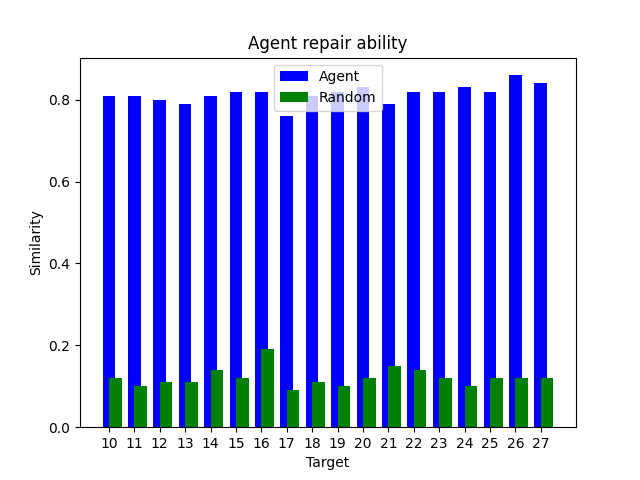}
    \caption{Ability of the agent to repair a noisy starting state to a Lego car.}
    \label{fig:lego_repair}
\end{figure}

\begin{figure}
    \centering
    \includegraphics[width=\linewidth]{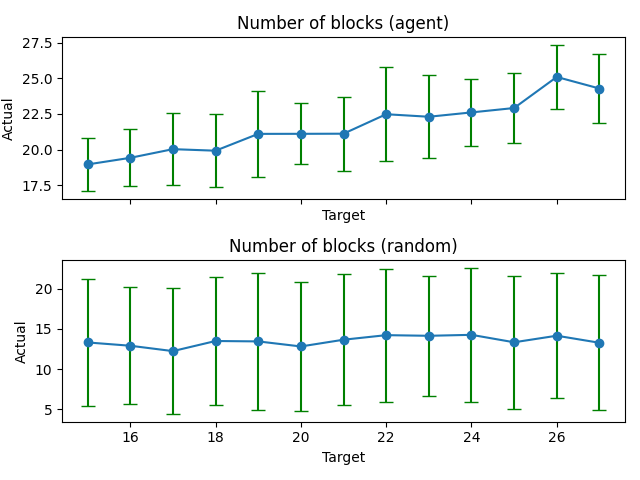}
    \caption{Controllability of Lego cars. The model learns to control the number of blocks it uses to build a car.}
    \label{fig:lego_ctrl}
\end{figure}

\begin{figure*}
    \centering
    \begin{subfigure}[t]{0.24\linewidth}
        \centering
        \includegraphics[width=\linewidth]{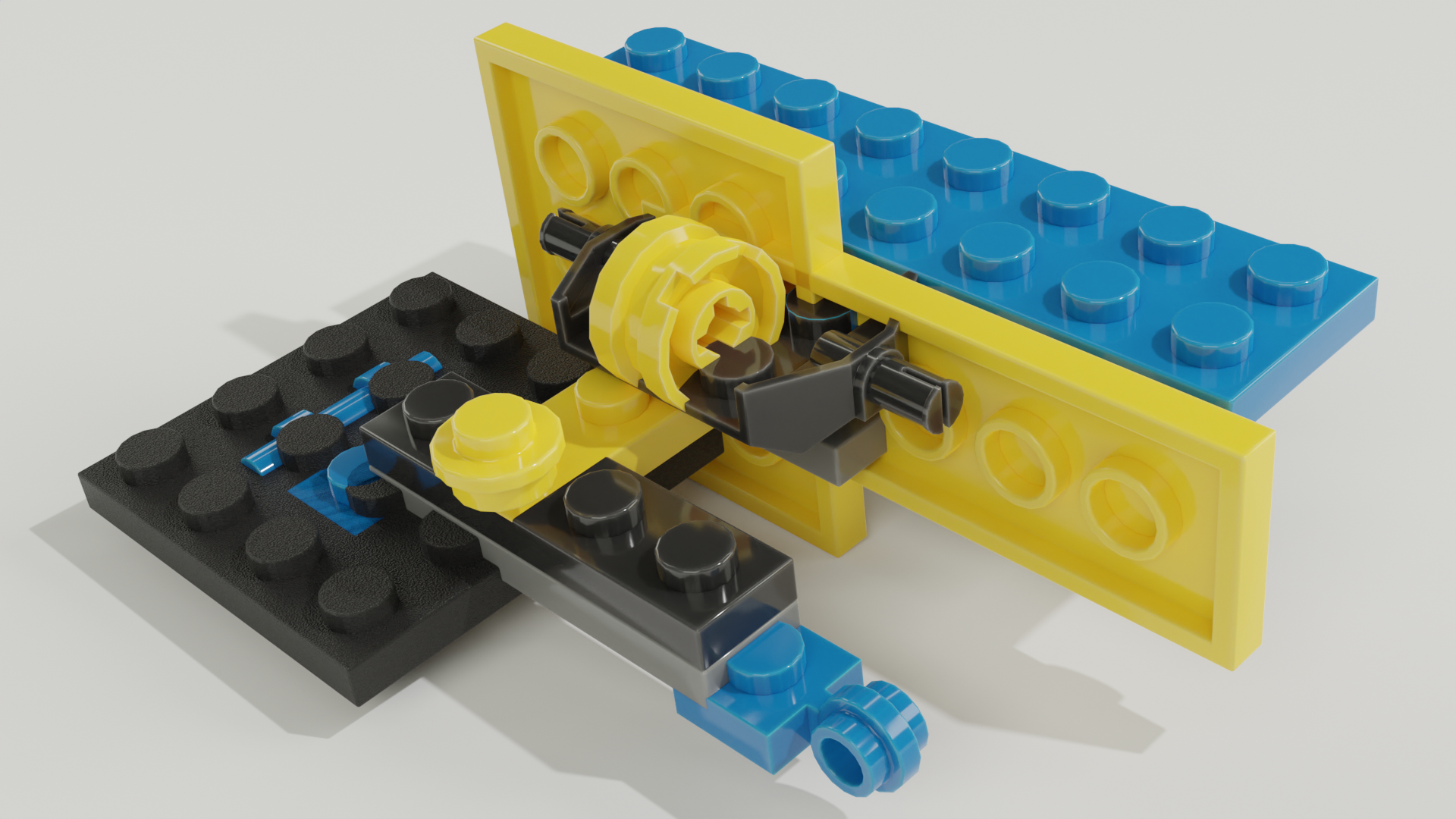}
        \caption{step 0}
        \includegraphics[width=\linewidth]{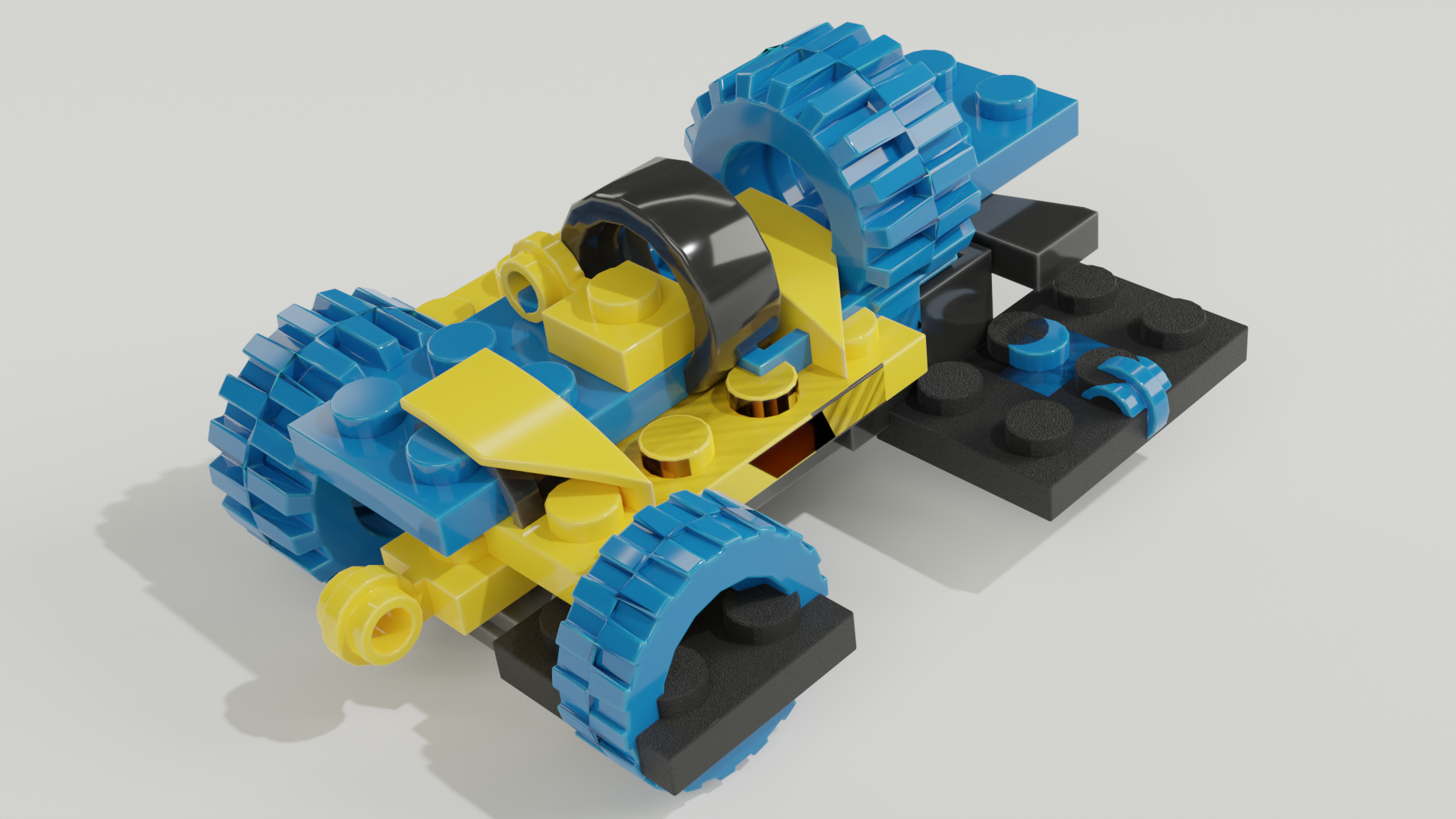}
        \caption{step 0}
    \end{subfigure}
    \begin{subfigure}[t]{0.24\linewidth}
        \includegraphics[width=\linewidth]{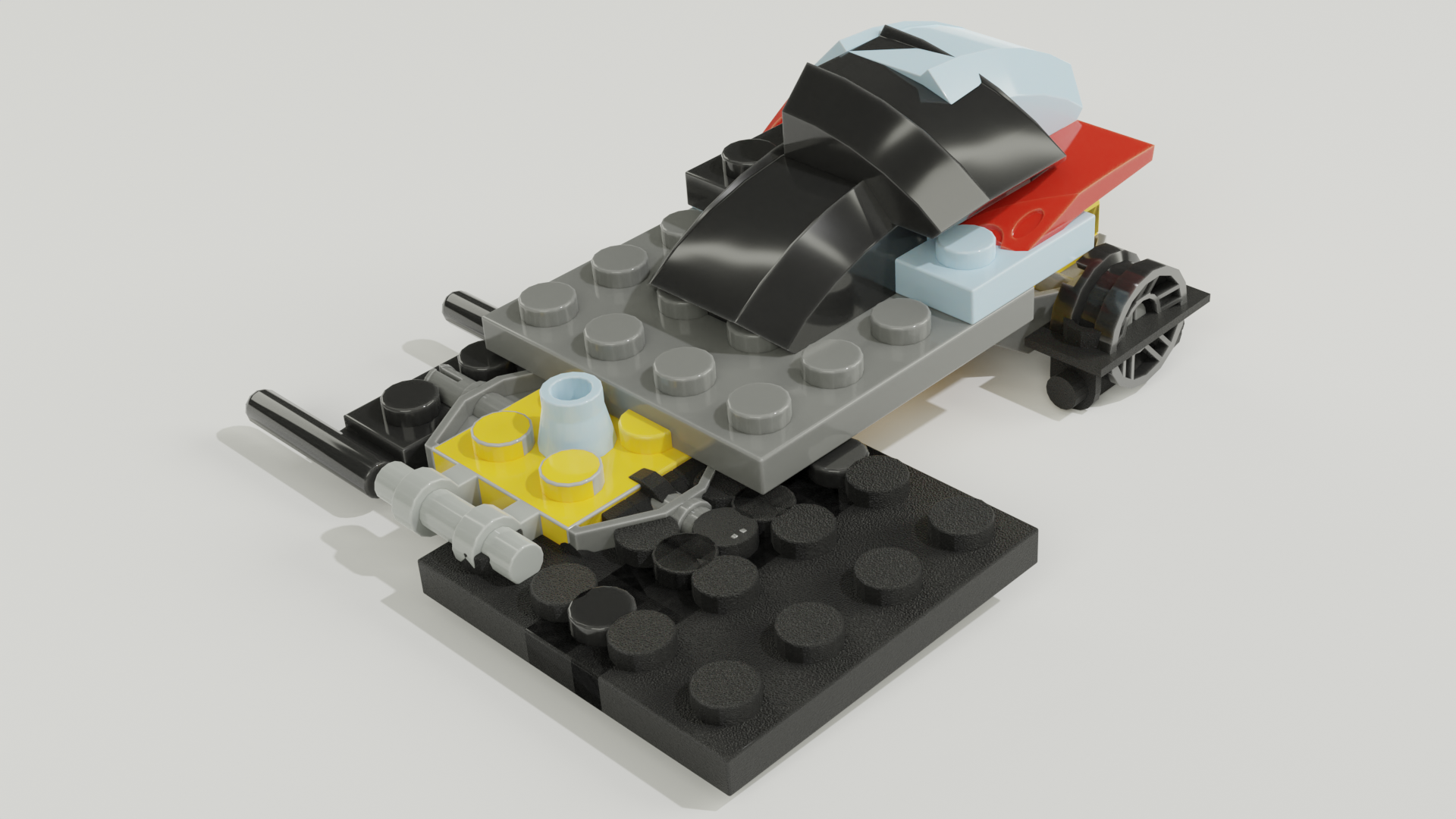}
        \caption{step 17}
        \includegraphics[width=\linewidth]{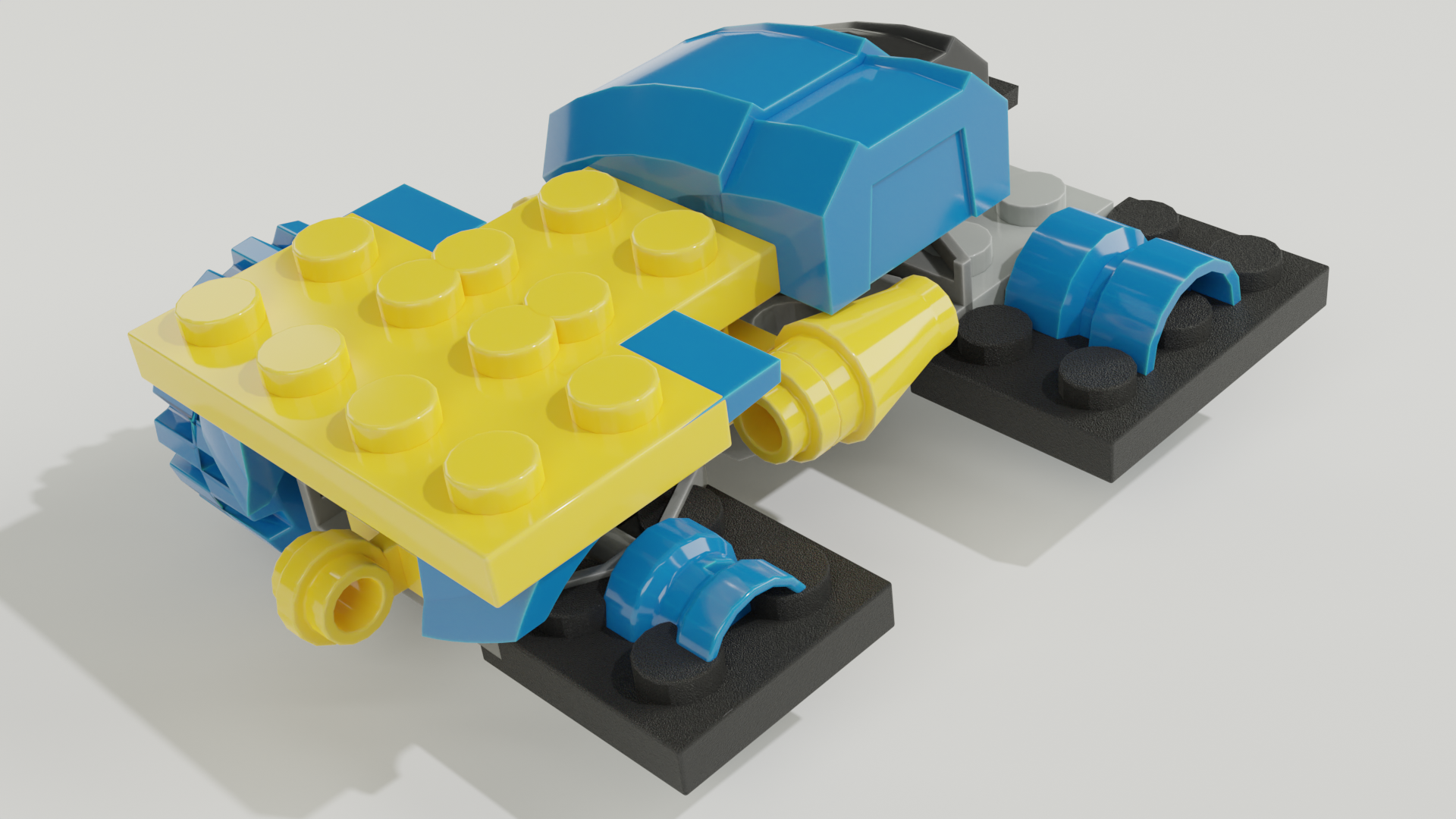}
        \caption{step 20}
    \end{subfigure}
    \begin{subfigure}[t]{0.24\linewidth}
        \includegraphics[width=\linewidth]{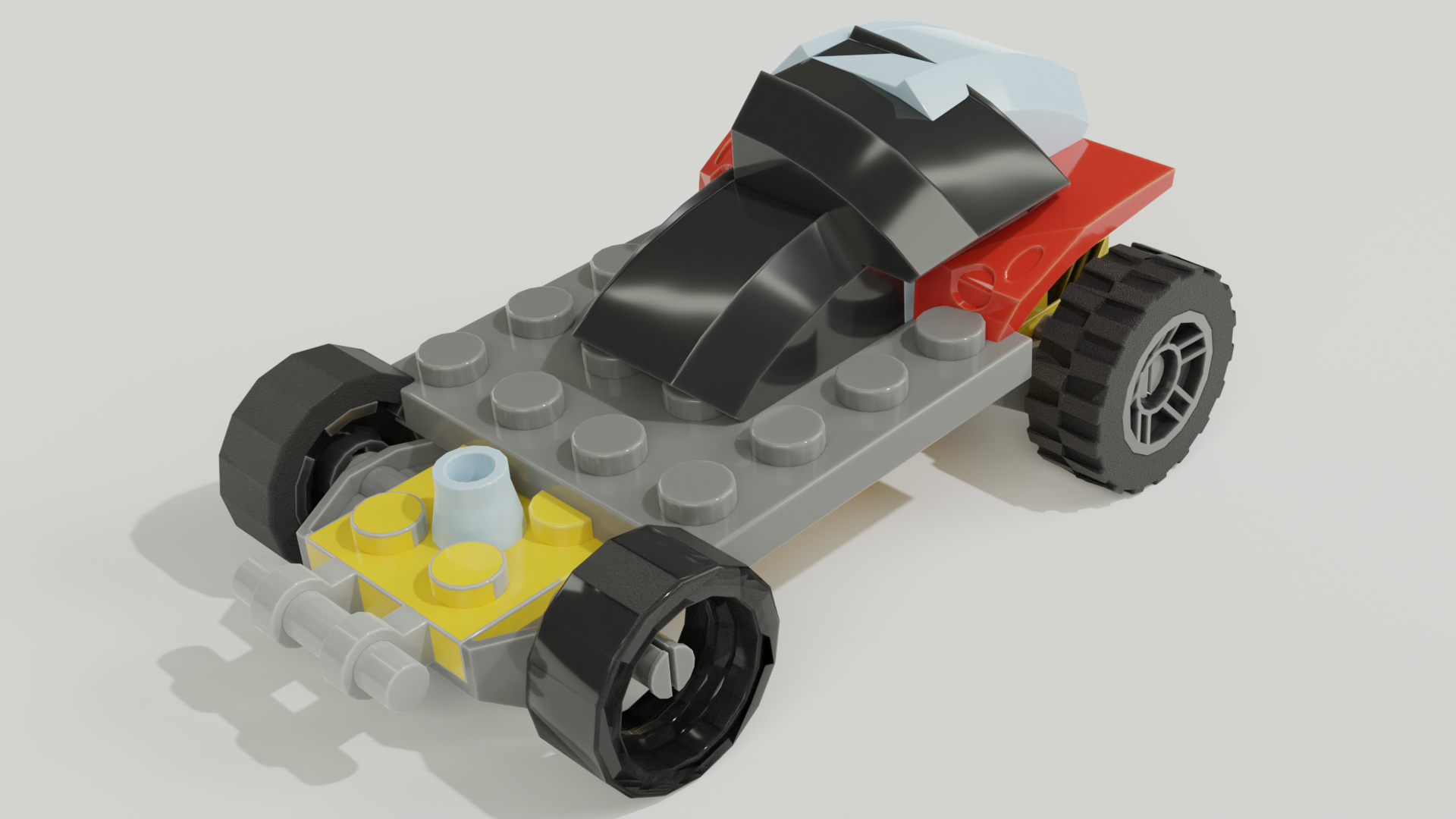}
        \caption{step 29}
        \includegraphics[width=\linewidth]{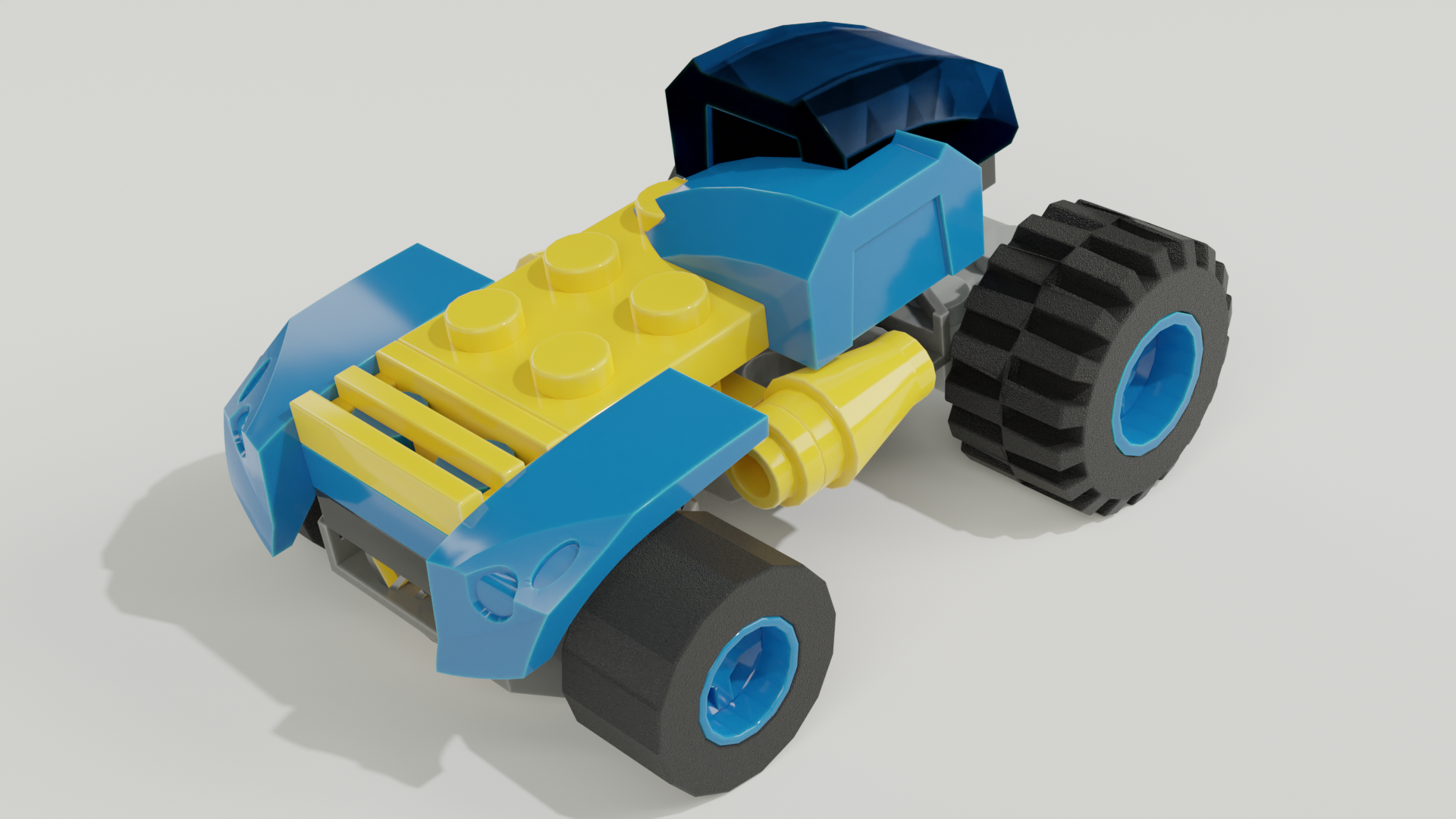}
        \caption{step 32}
    \end{subfigure}
    \begin{subfigure}[t]{0.24\linewidth}
        \includegraphics[width=\linewidth]{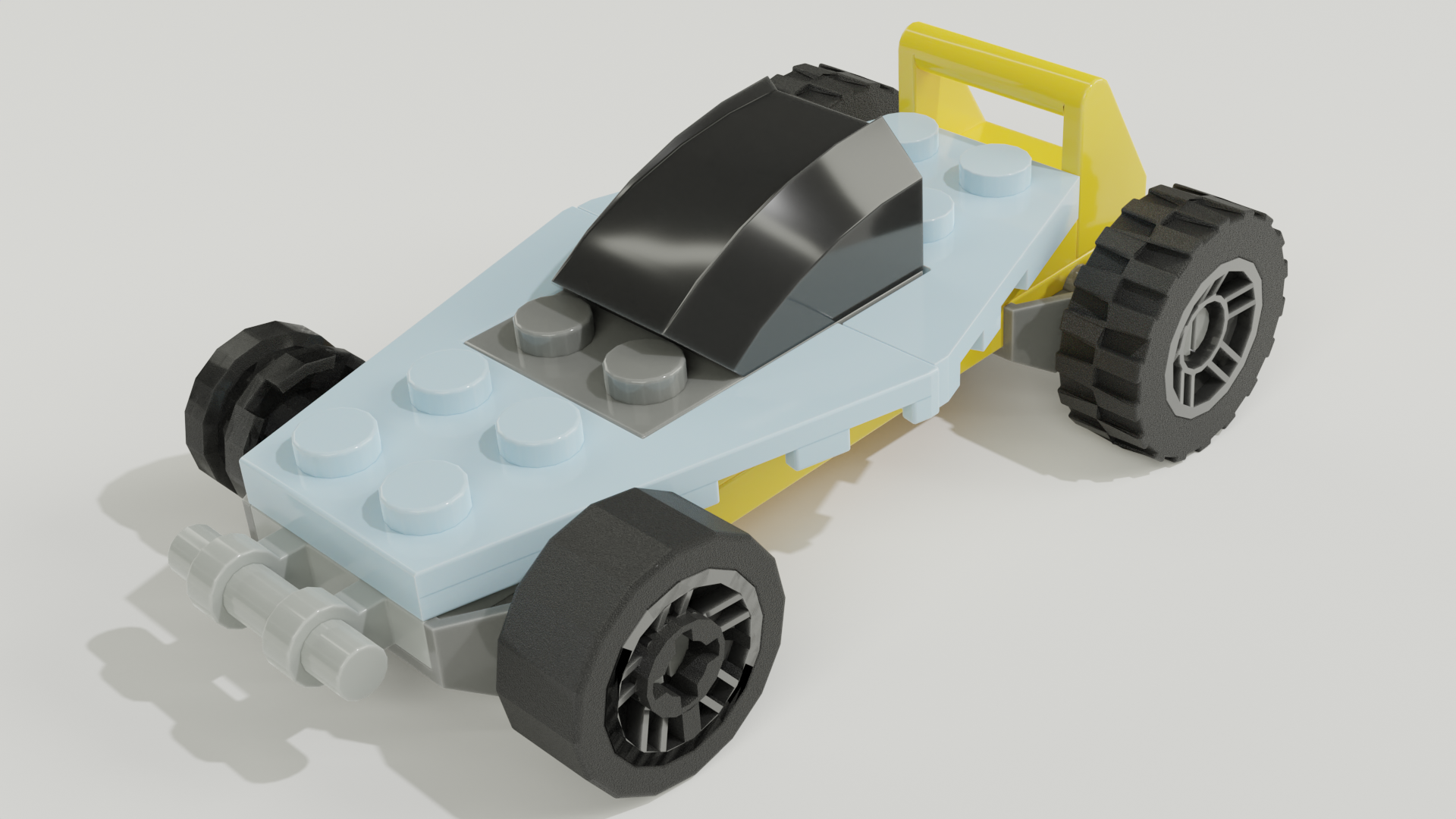}
        \caption{step 54}
        \includegraphics[width=\linewidth]{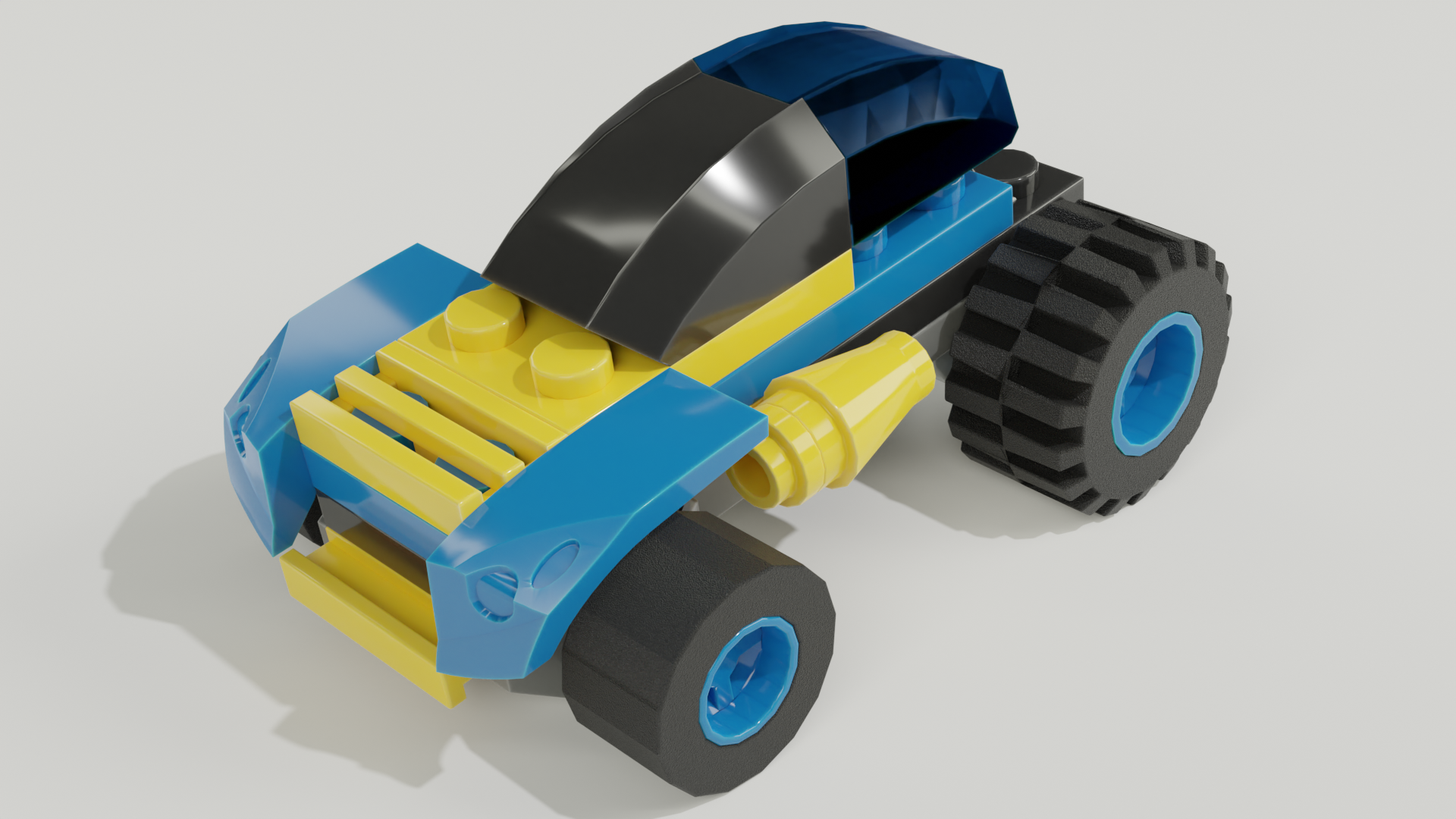}
        \caption{step 54}
    \end{subfigure}
    \caption{Inference episodes in which the agent repairs from a noisy starting state to a Lego car.}
    \label{fig:lego_example}
\end{figure*}

For the Lego experiments, we were concerned with whether an agent could learn (and how well) to repair noisy states to valid cars using a small dataset of 15 goal cars. Due to the recent creation of the ``legopcg'' framework, appropriate benchmark models aren't yet readily available. Therefore, we decided to compare the agent's capacity to repair and its controllability to random counterparts. Figure~\ref{fig:lego_repair} shows the average similarity of final states to the next closest goal for a given target metric. The blue bar, which indicates our agent shows that the agent is able to repair from noisy random states well across all of the target inputs tests, achieving around 80\% similarity to the next closest goal, respectively. By contrast, the random agent in green hovers between 10-20\% similarity which is competitive with the similarity of a given starting, indicating that the random agent makes no progress towards the goal. We also tested the generated cars to see how many of them have four wheels. This number was computed to be $23.91\% \pm 5.2$  over the 3 separate inference runs. This shows that the network was able to understand that to constitute a car requires the existence of 4 wheels.

Figure~\ref{fig:lego_ctrl} shows the controllability of the agent compared to its random counterpart. As is evident the red curve of the agent, while muted, is still pointing up and to the right. This is noticeably different than the random green curve which shows no directional trend up and to the right. We attributed the weakness in the controllability curve (as evidenced by the small domain) of our agent to the fact that we are not yet checking for a termination condition and instead fix the number of steps per episode. This problem can be solved either by having a neural network that can detect if the output is a Lego car or by training the network to know when it doesn't need to change. The second solution requires having a smart destructor that makes sure there is an equal distribution of all of the different condition parameters, including no-change. Figure~\ref{fig:lego_example} shows the repair steps towards generating two different cars that didn't exist in the original dataset. Midway through, the object starts resembling a car and the following steps were just polishing, suggesting a termination condition would help us generate some  interesting cars.

\section{Discussion}

One of the reasons diffusion models are so popular is their capability for a text-guided generation; this is how Stable Diffusion and Midjourney can give us high-quality images of so many things if we can only put them in words. The basic idea is to pass the prompt through a language model to get an embedding (a vector of some length, e.g. 512) and use this embedding as a conditional input in training and inference. In theory, the control scheme presented in this paper should be extendable to text-guided generation, so that we could ask the generator to generate a level or Lego creation according to any description. The big problem here is that given the large space of natural language prompts, having a large training set seems necessary, and this amount of training data is not necessarily available for the domains where PoD works best. However, this may be solvable by using a pre-training and fine-tuning scheme. Another solution to this problem is to try to use a smart destructor to generate different artifacts in the middle of destruction that can have different descriptions. This process could be automated using evolutionary algorithms and an image-to-language model such as CLIP~\cite{radford2021learning}.

While Path of Destruction and PCGRL differ greatly in what they are trained on -- existing content vs rewards -- they share a joint generation mode and network structure. This suggests that we could combine the two methods in various ways. For example, we could use Path of Destruction to generate levels and use PCGRL to repair or fix them in case they had errors. You could also imagine pre-training with PCGRL, perhaps over multiple different games, and fine-tuning with Path of Destruction, to make the best use of limited training data. Another idea could be using Path of Destruction as curriculum learning for PCGRL in the same manner as Justesen et al. work~\cite{justesen2018illuminating}. The PoD can be used to generate a destroyed level with a small trajectory of destruction where it is easy for PCGRL to repair it. When the PCGRL gets better at doing these repairs, we can increase the trajectory length until it reaches the whole level. This might sound the same as starting from complete noise and learning using PCGRL but in some hard domains, where the reward signal is sparse, we think this methodology will help overcome this problem.

\section{Conclusion}

We presented Controllable Path of Destruction, a new version of the Path of Destruction algorithm, and applied it to the generation of 2D Zelda levels and 3D Lego cars. Whereas the original Path of Destruction produced valid games conditioned only on noise (destroyed levels), this controllable version allows the user to guide the level-generation process toward relevant high-level features. This is achieved by annotating paths of random ``destructive'' edits with the features of each partially destroyed level state.  In both domains, the controllable Path of Destruction was able to generate proper content and maintain controllability with respect to most of the control parameters.

The Lego domain comprises the first application of Path of Destruction to 3D. We showed that with a small dataset of only 15 goal cars, the agent was able to learn to repair from a noisy starting state toward a goal, achieving an average of 80\% similarity to the closest goal car across all of the targets tested. Furthermore, we showed progress toward making the agent's controllability effective even though we didn't enforce any boundary thresholds on the input metric. Both of these constitute low-hanging opportunities that we believe should improve the controllability of self-supervised level generators. A lot of the issues raised by the new algorithm are due to the random destruction of the tiles. We hypothesize that using a smart destructor would elevate this problem by making sure that all the conditional input values are covered.

\bibliographystyle{IEEEtran}
\bibliography{ref}

\end{document}